\pdfoutput=1

\documentclass[11pt]{article}

\usepackage[final]{acl}

\usepackage{times}
\usepackage{latexsym}
\usepackage{comment}

\usepackage[T1]{fontenc}

\usepackage[utf8]{inputenc}

\usepackage{microtype}

\usepackage{inconsolata}

\usepackage{graphicx}
\usepackage{booktabs}

\usepackage{amsmath}
\usepackage{mdframed}
\usepackage{enumitem}

\usepackage{booktabs} 
\usepackage{amssymb}  

\title{Toward Human-Centered Readability Evaluation}

\author{
  \textbf{Bahar {\.I}lgen\textsuperscript{1}} \textnormal{and}
  \textbf{Georges Hattab\textsuperscript{1,2}}
\\
  \textsuperscript{1} Center for Artificial Intelligence in Public Health Research (ZKI-PH),\\Robert Koch Institute, Berlin, Germany\\
  \textsuperscript{2} Department of Mathematics and Computer Science,\\Freie Universität Berlin, Berlin, Germany
  \\ \{ilgenb, hattabg\}@rki.de
  }

\begin{document}
\maketitle
\begin{abstract}
Text simplification is essential for making public health information accessible to diverse populations, including those with limited health literacy. However, commonly used evaluation metrics in Natural Language Processing (NLP)—such as BLEU, FKGL, and SARI—mainly capture surface-level features and fail to account for human-centered qualities like clarity, trustworthiness, tone, cultural relevance, and actionability. This limitation is particularly critical in high-stakes health contexts, where communication must be not only simple but also usable, respectful, and trustworthy. To address this gap, we propose the Human-Centered Readability Score (HCRS), a five-dimensional evaluation framework grounded in Human-Computer Interaction (HCI) and health communication research. HCRS integrates automatic measures with structured human feedback to capture the relational and contextual aspects of readability. We outline the framework, discuss its integration into participatory evaluation workflows, and present a protocol for empirical validation. This work aims to advance the evaluation of health text simplification beyond surface metrics, enabling NLP systems that align more closely with diverse users’ needs, expectations, and lived experiences.
\end{abstract}

\section{Introduction}

 Text simplification — the process of replacing complex terms with simpler alternatives, removing extraneous details, or breaking down lengthy sentences while preserving essential meaning~\cite{chandrasekar1996motivations, saggion2017automatic} — is especially critical in health communication. Yet contemporary evaluation metrics often focus on the outcomes of individual simplification operations and fail to fully capture human judgments of overall simplicity. This gap is particularly concerning in the medical domain, where effective information delivery is essential for public well-being, especially for individuals with limited health literacy~\cite{nutbeam2000health, mccormack2013recommendations}. Health-related materials such as medication instructions, risk explanations, and care recommendations are often overly complex and cognitively demanding. Recent advances in natural language processing (NLP) have enabled automated systems to generate easier-to-read versions of such texts, aiming to improve accessibility and comprehension~\cite{siddharthan2014survey, espinosa-zaragoza-etal-2023-automatic, stajner-2021-automatic}.

 These systems are typically evaluated with automatic metrics such as SARI~\cite{xu2016optimizing}, FKGL~\cite{kincaid1975derivation}, or BLEU~\cite{papineni2002bleu}, which focus on surface‑level features (e.g., lexical simplicity, sentence length, or $n$-gram overlap). While useful for benchmarking, such metrics fail to capture whether simplified health information is genuinely clear, actionable, and trustworthy to real users—qualities that are critical in high‑stakes domains like public health. This challenge mirrors broader efforts in AI to align system performance with human values, where evaluation must go beyond automatic scores to incorporate human judgment. Reinforcement Learning from Human Feedback (RLHF)~\cite{10.1145/3743127} enables models to learn directly from user preferences and expert evaluations, serving as a gold standard for aligning system behavior with human values. However, collecting large-scale, high-quality human feedback can be costly and impractical. To address this limitation, recent approaches such as Reinforcement Learning with AI Feedback (RLAIF)~\cite{lee2024rlaifvsrlhfscaling} offer scalable alternatives to costly human evaluations, but risk misalignment with human values and needs in sensitive contexts.

To address the gap between metric‑driven evaluation and real‑world user experience, we investigate:
\begin{enumerate}[label=\textbf{Q\arabic*}]
    \item Which existing readability metrics (lexical, syntactic, semantic, or neural) align most strongly with human‑centered dimensions—clarity, trustworthiness, tone appropriateness, cultural relevance, and actionability—in simplified health texts?
    \item To what extent can a composite Human‑Centered Readability Score (HCRS), integrating automatic features with human feedback, achieve stronger alignment with user evaluations than the best standalone metric? 
    \item How can human–computer interaction (HCI) techniques, such as interactive feedback collection and participatory design, be embedded into the evaluation pipeline to make metric optimization more responsive for real‑world needs?
\end{enumerate}

Readability in health communication is inherently multidimensional: a linguistically simple message can still be unclear, culturally inappropriate, or untrustworthy. Addressing these dimensions requires going beyond surface‑level metrics. Our contributions are threefold:
\begin{itemize}
    \item \textbf{Redefinition:} Reconceptualizing readability for health text simplification from a human‑centered perspective, drawing on HCI and health communication research.
    \item \textbf{Framework:} Proposing a five‑dimension conceptual model—clarity, trustworthiness, tone appropriateness, cultural relevance, and actionability—that can guide evaluation and system design.
    \item \textbf{Agenda:} Arguing that these dimensions must inform both the evaluation and design of simplification systems, especially for vulnerable or marginalized audiences, and outlining a future research agenda bridging NLP and HCI.
\end{itemize}

By shifting the evaluation lens from system‑centric to user‑centric, we aim to advance health‑focused NLP systems that are not only technically effective, but also socially and culturally responsive to the needs of diverse real‑world users—an imperative in public health communication.

\section{Background: Readability in NLP and Health Communication}
\subsection{Automatic Readability Metrics in NLP}
Readability in NLP is often reduced to numerical scores, yet human perception of clarity, trust, and usability depends on far more than surface form. Automatic evaluation of text simplification in NLP typically falls into three categories: 
 \emph{surface-level metrics} (e.g., BLEU, SARI), \emph{semantic metrics} (e.g., BERTScore, QuestEval, METEOR), and \emph{readability indices} (e.g., FKGL, SMOG, Coleman–Liau), the latter often applied in health communication. 

 \textbf{BLEU}~\cite{papineni2002bleu} computes $n$-gram precision between outputs and references, penalizing brevity; \textbf{SARI} \cite{xu2016optimizing} measures the quality of $n$-gram ``keep'', ``add'', and ``delete'' operations; \textbf{FKGL} computes a linear combination of average sentence length and syllables per word to estimate grade level. Despite their widespread use in NLP benchmarking, these metrics neglect cognitive, emotional, and social dimensions central to human perception of readability—factors that are critical in high-stakes settings such as public health.

\noindent\framebox[\linewidth][l]{%
\parbox{0.97\linewidth}{\small
\textbf{SARI:} $SARI = \frac{1}{3}(F_{add} + F_{keep} + F_{del})$ \\[0.5ex]
\textbf{BLEU:} $BLEU = BP \cdot \exp\left(\sum_{n=1}^{N} w_n \log p_n\right)$ \\[0.5ex]
\textbf{FKGL:} $FKGL = 0.39 \times \frac{\text{Words}}{\text{Sentences}} + 11.8 \times \frac{\text{Syllables}}{\text{Words}} - 15.59$
}%
}

Recent neural and reference-based metrics, such as \textbf{BERTScore}~\cite{zhang2020bertscore} (contextualized token similarity) and \textbf{QuestEval}~\cite{scialom2021questeval} (Q\&A-based semantic evaluation), and \textbf{SALSA}~\cite{heineman-etal-2023-dancing} (edit-based granular evaluation), aim to capture meaning and content transformation beyond surface forms. SALSA, in particular, provides a fine-grained typology of simplification edits and an automatic variant (LENS-SALSA) to score outputs without references, showing stronger alignment with human judgments in recent studies. While SALSA covers simplification edits in detail, it primarily captures structural and lexical changes and cannot fully assess whether the resulting text is clear to diverse users in real contexts. Yet, they still fail to assess \emph{trustworthiness}, \emph{emotional resonance}, or \emph{context appropriateness}—qualities that determine whether a health message is actually usable. These limitations are also evident in large‑scale meta‑evaluations. For example, Alva-Manchego et al.~(\citeyear{alva2021unsuitability}) found that these metrics often correlate only weakly with human judgments, especially for multi‑operation simplifications where lexical, syntactic, and semantic changes co‑occur. They systematically compared leading metrics across several simplification systems and found that, while BERTScore Precision achieved the highest overall correlation, performance dropped sharply for high-quality outputs, underscoring the persistent gap between metric scores and user-perceived quality. As shown in Table~\ref{tab:metric_coverage}, no widely used automatic metric provides comprehensive coverage of the user‑centered dimensions that are critical in health contexts—namely clarity, trustworthiness, and actionability.

\begin{table*}[h!]
\centering
\footnotesize
\setlength{\tabcolsep}{3pt}
\begin{tabular}{lcccccc}
\toprule
\textbf{Metric} & \textbf{Lexical Simplicity} & \textbf{Syntactic Simplicity} & \textbf{Semantic Adequacy} & \textbf{Clarity} & \textbf{Trustworthiness} & \textbf{Actionability} \\
\midrule
BLEU & \checkmark & (\checkmark) & -- & -- & -- & -- \\
SARI & \checkmark & (\checkmark) & -- & -- & -- & -- \\
FKGL & \checkmark & \checkmark & -- & $(\checkmark^{1})$ & -- & -- \\
BERTScore & -- & -- & \checkmark & -- & -- & -- \\
QuestEval & -- & -- & \checkmark & -- & -- & -- \\
SALSA & \checkmark & \checkmark & -- & $(\checkmark^{2})$ & -- & -- \\
\bottomrule
\end{tabular}
\caption{Coverage of key readability and user-centered dimensions by common automatic metrics. (\checkmark): partially addresses; --: not addressed. 
\textsuperscript{1} FKGL partially reflects clarity by rewarding shorter sentences and simpler words, but does not capture semantic or user-perceived clarity. \textsuperscript{2} SALSA contributes to clarity by typologizing lexical and structural simplification edits, but does not directly assess whether outputs are clear to diverse users.}
\label{tab:metric_coverage}
\end{table*}

\subsection{Human Perception and the Readability Gap}
Prior work in HCI and health communication \cite{ishikawa2010health, crossley2016taaco} underscores that readability is constructed through users' prior knowledge, affect, and social context, as well as deeper textual properties such as local and global cohesion. Consequently, leading researchers now advocate for \emph{mixed-method} evaluation frameworks combining robust automatic metrics with validated user scales (e.g., Trust in Health Information Questionnaire, cognitive load indices) and real-time feedback. While such calls for mixed-method evaluation are compelling, there remains limited empirical evidence demonstrating how current metrics diverge from human perception in practice.

Empirical evidence confirms the gap between automatic scores and lived experience. In a controlled study, \cite{leroy2022evaluation} found that simplified health texts significantly improved comprehension accuracy—boosting correct recall from 33\% to 59\%—with the largest gains among participants with lower education or limited English proficiency. Earlier work \cite{leroy2013user} similarly reported improvements in perceived clarity and learning outcomes. Taken together, these studies illustrate that surface-level gains in simplicity do not necessarily ensure that health information is perceived as accessible or actionable by diverse users. However, more recent NLP-based evaluations (e.g., \cite{alva2021unsuitability, maddela-etal-2021-controllable}) show that even high-SARI outputs can be perceived as emotionally flat or insufficiently actionable, suggesting that surface simplicity does not guarantee effective communication.

In public health, \emph{readability is relational}. A text must be clear, but also trustworthy, respectful, culturally sensitive, and actionable. This gap is both methodological and conceptual: current NLP evaluation practices rarely capture these interpersonal and contextual dimensions. For example, a jargon-free vaccine information sheet delivered in an emotionally cold or overly directive tone can still alienate its audience. These interpersonal and contextual factors are invisible to current NLP metrics, which rarely incorporate user studies or grounded sociocultural analysis.

This disconnect is consequential: in health contexts, misunderstanding can reduce adherence, increase anxiety, or even cause harm. We argue that evaluating readability in NLP must center on \emph{user interpretation}—not just algorithmic scores—requiring new, multidimensional metrics co-designed with end users. In short, a truly ``readable'' health text is one that is \emph{understood, trusted, respected, and acted upon}. Achieving this requires moving beyond current benchmarks toward evaluation frameworks that are co‑designed with and validated by target user communities.

\section{Limitations of Current Metrics}

Despite widespread adoption, automatic evaluation metrics for text simplification—such as SARI, FKGL, and BLEU—face critical shortcomings in public health communication.

\paragraph{Intrinsic Metric Limitations.}
These metrics are designed to capture primarily surface‑level transformations—such as lexical substitution, sentence compression, and $n$‑gram overlap with reference texts—and are effective for system‑level benchmarking. However, they provide no direct evidence regarding whether a simplified text is, in fact, clear, trustworthy, or pragmatically usable for its intended audience.

\paragraph{Overreliance on Reference-Based Evaluation.}
One major limitation is the overreliance on reference-based evaluation. Metrics like BLEU and SARI compare system outputs to one or more human-written references, assuming that overlap with these references implies higher quality. However, simplification is an inherently subjective task with high variance in valid outputs. A single sentence can be simplified in many plausible ways depending on the user's background knowledge, cultural expectations, or even emotional state. Penalizing deviations from a limited reference set risks excluding useful and contextually appropriate simplifications.

\paragraph{Neglect of Pragmatic and Relational Aspects.}
Another concern is the neglect of pragmatic and relational aspects of language use. Metrics such as FKGL reduce readability to sentence length and syllable count, ignoring tone, politeness, cultural sensitivity, and actionability. In health contexts, these dimensions are critical. A sentence that is syntactically simple but emotionally flat or overly authoritative may alienate readers or diminish trust in the message. Beyond these pragmatic concerns, relational aspects such as trust-building, respect, and perceived empathy are equally essential, yet remain invisible to current metrics, which operate without understanding user intent, affective response, or context of use.

\paragraph{Limited Generalization Across User Populations.}
Moreover, automatic metrics do not generalize well across diverse user populations. Individuals with different literacy levels, cultural backgrounds, or health conditions may interpret the same text in divergent ways. Metrics grounded in average-case assumptions fail to capture these differences, potentially reinforcing existing disparities in health communication. For example, a simplification that seems effective for English-dominant, college-educated users may confuse or offend speakers of other dialects or individuals with lower health literacy.

\paragraph{Risks from AI-Generated Feedback Loops.}
Finally, the increasing use of reinforcement learning with AI-generated feedback (RLAIF) raises additional concerns. When optimization is driven by synthetic evaluators trained on limited metrics or preferences, system behavior may diverge from human-centered values. Without human-in-the-loop validation, systems risk overfitting to numerical proxies that do not reflect lived experience or real-world comprehension.

\paragraph{Evidence from Recent Studies}
Recent research ~\cite{choi2024combining} shows that dynamically combining multiple evaluation metrics (lexical and semantic) yields much stronger alignment with human ratings than any single metric can. Large-scale meta-evaluations ~\cite{alva2021unsuitability} found that commonly used automatic metrics, such as BLEU and SARI, typically exhibit only low-to-moderate correlation with human judgments on simplification quality—particularly when multiple rewriting operations are involved. These results underscore the risk of relying solely on surface-level scores, as high metric values may not reflect true gains in user-perceived clarity or simplicity.

\medskip
Overall, these limitations highlight the need for evaluation frameworks that move beyond narrow linguistic proxies to capture the relational, pragmatic, and affective dimensions of readability in health-focused NLP. Building on this,  we argue for extending ensemble approaches with participatory user evaluation to address these gaps.

\section{Human-Centered Readability Score (HCRS)}

\begin{figure}[ht]
  \centering
  \includegraphics[width=1\linewidth]{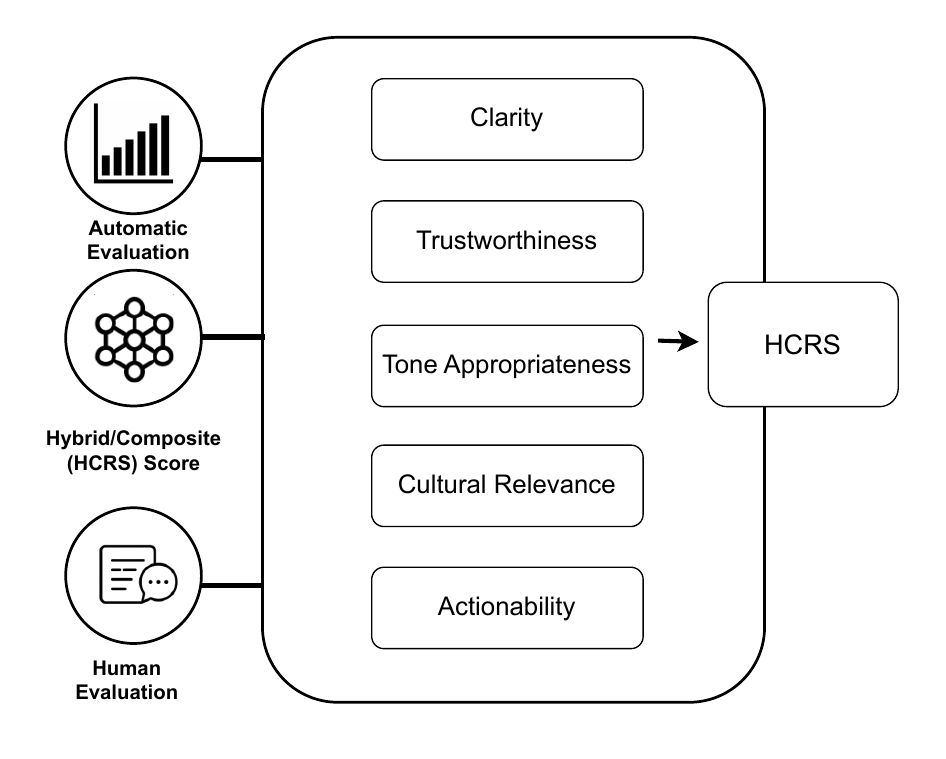} 
  \caption{HCRS framework diagram}
    \label{fig:hcrs}
\end{figure}

We argue that readability in health communication should be redefined not as a property of the text alone, but as a dynamic, relational, and context‑sensitive experience shaped by human perception. To address the limitations of current surface‑level metrics, we introduce a five‑dimensional framework—\textit{clarity, trustworthiness, tone appropriateness, cultural relevance, and actionability}—grounded in research from health communication, HCI, and participatory design. Each dimension captures a distinct yet interdependent aspect of how users interpret and engage with simplified content. Figure~\ref{fig:hcrs} illustrates the five core dimensions and the integration of automatic and human evaluation within the hybrid HCRS protocol.

\subsection{Clarity}
The core dimension of readability is whether a text can be readily understood by its intended audience. This involves removing jargon, reducing syntactic complexity, and ensuring a logical flow of ideas. Crucially, clarity must be defined from the user’s perspective: a linguistically simple sentence can still be unclear if it lacks context, relies on unfamiliar metaphors, or omits essential background information. In health communication, clarity should be measured not solely by word and sentence length, but by the degree to which users can accurately and confidently extract the intended meaning. Clarity measurement is automatically scored via readability indices (FKGL, SMOG), jargon detection, and cohesion tools (e.g., Coh-Metrix~\cite{Graesser2004}), combined with user-rated comprehension and ease-of-reading survey items on a 5-point Likert scale or direct comprehension quizzes.

\subsection{Trustworthiness}
In the context of health communication, \emph{trustworthiness} refers to the perceived reliability, credibility, and transparency of the source, distinct from the broader and often contested notion of “trust” in general AI ethics debates. Health-related texts are not merely informational—they are also relational. Users often evaluate not just what is said, but who is saying it and how it is said. Trust is influenced by perceived author credibility, transparency, and tone. Simplified texts that are overly generic, impersonal, or dismissive of complexity may erode trust, particularly among marginalized populations with historical reasons to be skeptical of medical authority. A readable text should convey not only facts, but also empathy and accountability. Broader debates on “trust” in AI are addressed in the Discussion, but here the focus remains on relational trust in patient-facing health communication. 

Trustworthiness is quantified by detecting explicit source attribution, institutional language, transparency features, and domain authority markers in the text; supplemented by participant ratings of perceived credibility, transparency, and author reliability gathered from validated survey instruments.

\subsection{Tone Appropriateness}
The emotional tone of a message can profoundly affect how it is received~\cite{Street2009CommunicationHeal,Hinyard2007, Arora2003}. In simplified texts, tone often shifts unintentionally—becoming condescending, overly directive, or emotionally flat. Especially in health contexts, tone must balance clarity with compassion, authority with humility. An appropriate tone affirms the reader’s dignity, avoids blame, and fosters a sense of collaboration rather than control.
Tone appropriateness is defined as the alignment of a text’s affective, pragmatic, and interpersonal cues with the expectations and sensitivities of the target user group, especially in health communication contexts. Technically, tone can be operationalized using multidimensional features derived from both automatic computational models and structured human assessments:

\subsubsection*{Automatic Assessment}
\begin{itemize}
    \item \textbf{Pragmatic Feature Extraction:} Quantify politeness strategies using established classifiers (e.g., Stanford Politeness Classifier, formality index). 
    Key features include indirectness, mitigation (e.g., ``could you'' instead of imperatives), and hedging (e.g., ``perhaps'', ``might'')~\cite{DanescuNiculescuMizil2013}.
    
    \item \textbf{Sentiment and Emotion Analysis:} Compute sentiment polarity (positive, neutral, negative) and emotional valence using transformer-based models (e.g., BERT, RoBERTa)~\cite{Devlin2019, liu2019roberta, MohammadTurney2013_Crowdsourcing} fine-tuned for affective state detection.
    
    \item \textbf{Empathy \& Support Classifiers:} Use models trained to detect empathy, warmth, and nonjudgmental language (e.g., EmpathBERT) to assess supportive tone~\cite{sharma-etal-2020-computational, guda-etal-2021-empathbert}.
    
    \item \textbf{Lexical Diversity \& Intensity:} Calculate frequencies of intensifiers (e.g., ``very'', ``extremely''), modals (e.g., ``should'', ``must''), evidentials (e.g., ``it seems''), and negative polarity items (e.g., ``never'', ``cannot'') that may indicate directive or controlling language~\cite{Biber1999}.
\end{itemize}

\subsubsection*{Human-Centered Evaluation}
\begin{itemize}
    \item \textbf{Likert-Scale Survey Items:} Collect ratings on standardized questions such as:
    ``This message feels respectful and supportive.'' or
    ``The tone is appropriate for the intended audience,'' following validated health communication scales.
    
    \item \textbf{Annotation Protocols:} Engage expert or target population annotators using detailed codebooks specifying tone-related phenomena (e.g., respectfulness, blame avoidance, collaborative framing) to ensure consistent evaluation.
\end{itemize}

The tone appropriateness for each text can then be computed as a weighted hybrid score:

\begin{equation}
\mathrm{Tone}_{\mathrm{HCRS}} =
\alpha_{1} P_{\mathrm{a}} +
\alpha_{2} S_{\mathrm{a}} +
\alpha_{3} E_{\mathrm{a}} +
\beta H
\end{equation}

\noindent
where $P_{\mathrm{a}}$: politeness score (auto), 
$S_{\mathrm{a}}$: sentiment score (auto), 
$E_{\mathrm{a}}$: empathy score (auto), 
$H$: human Likert rating and the $\alpha$ and $\beta$ coefficients are determined empirically via calibration on validation data to maximize alignment with user perceptions.

\subsection{Cultural Relevance}
Cultural relevance refers to the extent to which a simplified text preserves, reflects, and respects the cultural, linguistic, and social norms of its intended audience~\cite{resnicow1999cultural, KreuterMcClure2004, osborne2006health}. Texts can embed cultural meaning through references, metaphors, idioms, visual symbols, and formatting conventions; these elements may facilitate comprehension for in‑group readers but create barriers for out‑group readers.

From an evaluation standpoint, cultural relevance can be operationalized via (i) \emph{automatic detection} of culturally specific lexical items, named entities, and idiomatic expressions, combined with cross‑linguistic/multilingual embedding similarity measures to check alignment with target‑culture corpora; and (ii) \emph{human assessment} using Likert‑scale items measuring perceived familiarity, inclusivity, and absence of culturally alienating content. Loss of culturally meaningful content or introduction of inappropriate cultural markers during simplification can reduce both accessibility and trust. 
Accordingly, incorporating cultural‐perspective checks into simplification system design and evaluation is essential to ensure inclusivity and equity.

{\small
\begin{align}
\mathrm{Culture}_{\mathrm{HCRS}} &= 
\gamma_{1} \cdot E_a \;+\;
\gamma_{2} \cdot I_a \nonumber \\
&\quad+\; \gamma_{3} \cdot M_a
\;+\; \delta \cdot H_h
\end{align}
}

\noindent\textbf{where:} 
$E_a$: automatic entity match score (NER-based), 
$I_a$: automatic idiom/cultural expression match score, 
$M_a$: automatic multilingual embedding similarity score, 
$H_h$: human-rated cultural relevance (Likert), 
$\gamma_{1},\gamma_{2},\gamma_{3},\delta$: weights calibrated on validation set.

\subsection{Actionability}
In the final dimension, we address the need for simplified health texts to support informed action. This includes not only understanding a message, but knowing what steps to take and feeling empowered to take them. Actionability requires that information be specific, time-relevant, and contextually grounded in the user’s lived reality. Inadequate or ambiguous directives—e.g., “seek care if needed”—can confuse rather than guide. A readable text should reduce cognitive load while increasing behavioral clarity. 

From an evaluation perspective, actionability can be assessed via (i) automatic analysis of directive and procedural language (imperatives, explicit instructions, temporal/agent specification) and (ii) human ratings on validated Likert‑scale items capturing perceived clarity of next steps. These scores can be integrated into a hybrid metric within the HCRS framework~\cite{vishnevetsky2018interrater, KreuterMcClure2004}.

Operationally, actionability can be measured through (i) \emph{automatic linguistic analysis}, such as the detection and scoring of imperative verbs, procedural language, and explicit guidance framing; and (ii) \emph{human rating} via Likert-scale items assessing whether the reader feels well-informed and able to act on the information provided.

{\small
\begin{align}
\mathrm{Action}_{\mathrm{HCRS}} &= 
\lambda_{1} \cdot D_a \;+\;
\lambda_{2} \cdot P_a \nonumber \\
&\quad+\; \lambda_{3} \cdot Q_a
\;+\; \mu \cdot H_h
\end{align}
}

\noindent\textbf{where:} \\
$D_a$: automatic directive language/imperative score,\\
$P_a$: automatic procedural/instruction cue score,\\
$Q_a$: automatic presence of action-associated entities (temporal, agent, location),\\
$H_h$: human rating of perceived actionability (Likert),\\
$\lambda_{1},\lambda_{2},\lambda_{3},\mu$: weights calibrated on validation set.

By framing readability as a multidimensional construct, we aim to support the development of NLP systems that are better aligned with human needs and social context. Each of these dimensions is measurable, at least in principle, through user-centered evaluation methods such as interviews, surveys, and participatory testing. Importantly, these dimensions are not intended to be exhaustive or mutually exclusive, but to offer a starting point for rethinking evaluation as an interdisciplinary, collaborative process. 

\begin{figure}[ht]
    \centering
    \includegraphics[width=0.4\textwidth]{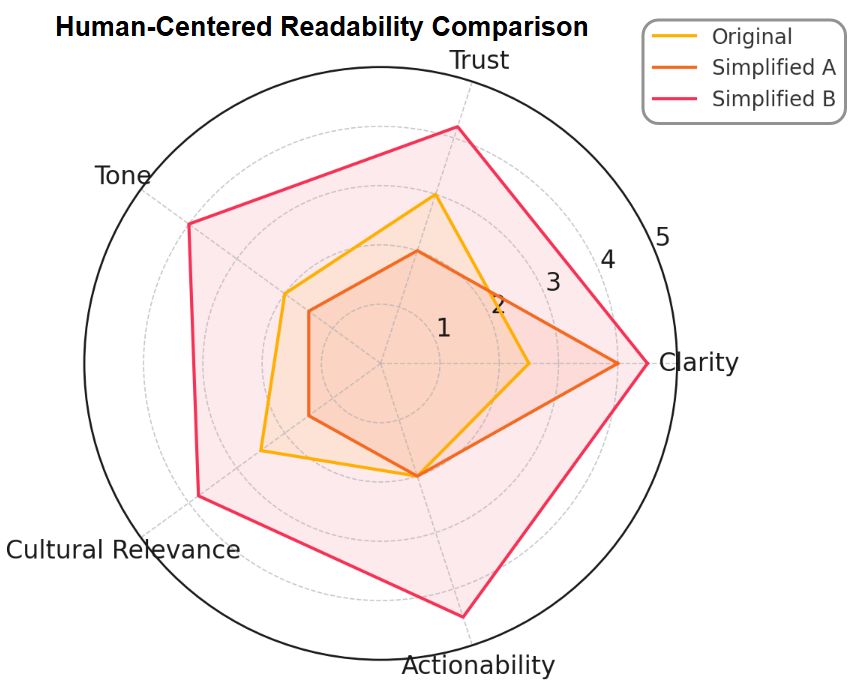}
    \caption{Illustrative comparison of original and simplified versions across five human-centered readability dimensions. Simplified B scores higher on trust, tone, and actionability, reflecting better alignment with user-centered design principles.}
     \label{fig:comparison}
\end{figure}

\begin{figure*}[ht]
  \centering
  \includegraphics[width=\textwidth]{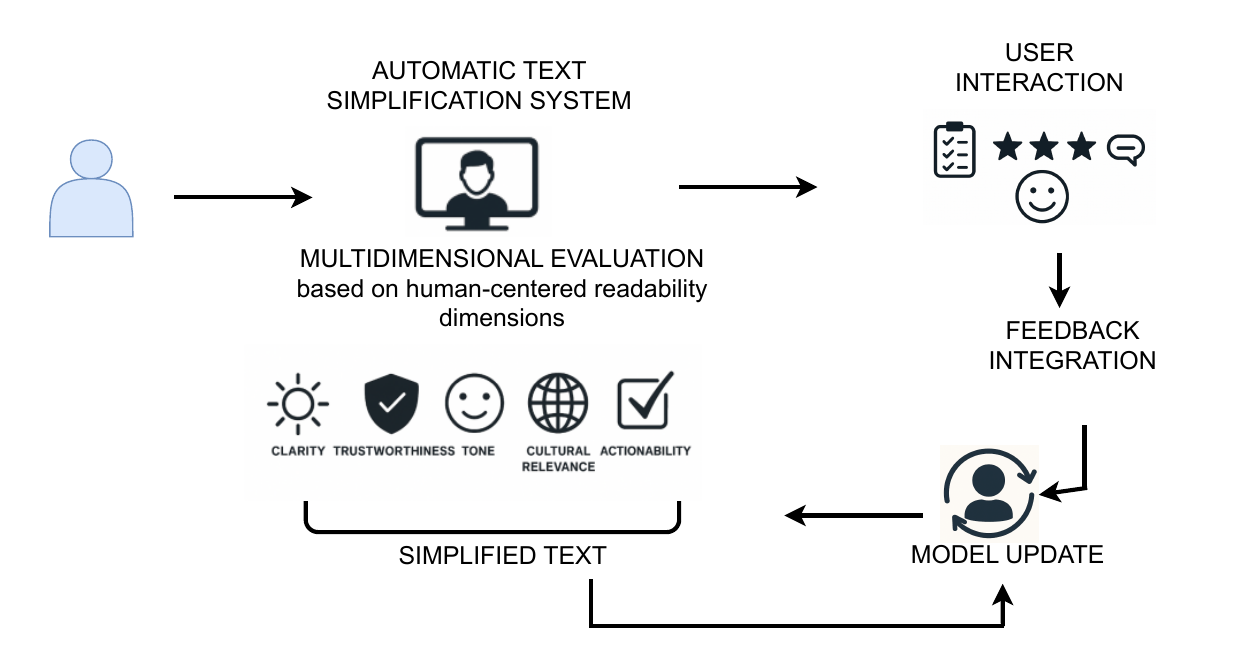}
  \caption{Human-in-the-loop readability evaluation}
  \label{fig:hitl}
\end{figure*}

Figures~\ref{fig:comparison} and~\ref{fig:hitl} illustrate the HCRS framework in practice. Figure~\ref{fig:comparison} compares original and simplified health texts across the five human‑centered dimensions, with version B scoring higher in trust, tone, and actionability. 
Figure~\ref{fig:hitl} shows how such scores can be integrated into a human‑in‑the‑loop evaluation pipeline, linking user feedback to model updates.

\section {Empirical Evaluation \& Discussion}
\label{sec:planned-eval}
This work defines a comprehensive experimental protocol to validate the Human-Centered Readability Score (HCRS) across diverse user populations and health literacy levels. The protocol integrates automatic and human-centered measures to move evaluation beyond current surface-oriented benchmarks.

\paragraph{Interactive \& Participatory Evaluation.}
Most NLP evaluation workflows are system-facing. To operationalize HCRS dimensions such as tone, trust, and actionability, user-facing interfaces should collect real-time feedback. Short prompts—e.g., “Was this sentence clear to you?”, “Did you feel respected by the way it was phrased?”—can be embedded in studies or deployment. Inclusive design, informed by participatory methods, engages diverse stakeholders, co-creates evaluation criteria, and adapts interfaces to cultural and literacy contexts. Accessibility features (visual icons, read-aloud options) can broaden participation, and lightweight web tools or crowdsourcing can scale data collection.

In practice, participatory methods can be embedded in evaluation pipelines through lightweight annotation interfaces where end-users quickly rate sentences on the five HCRS dimensions. Feedback can be collected in short micro-surveys (5–10 minutes), aggregated, and then reviewed with stakeholders (e.g., patients, clinicians, or domain experts) during participatory workshops. This creates an iterative loop: (i) immediate micro-ratings during evaluation, (ii) stakeholder workshops to review and refine criteria, and (iii) iterative updates where refinements directly inform HCRS weighting. For instance, in a vaccine information use-case, patients could rate clarity and trustworthiness via inline sliders, while health professionals review aggregated outputs to recalibrate actionability guidelines. This illustrates how participatory methods can be concretely integrated rather than remaining abstract, serving as a proof-of-concept protocol that demonstrates how HCRS could be operationalized through lightweight user studies even before large-scale deployment.

\paragraph{Structured, Multidimensional Feedback.}
Binary ratings rarely explain why a sentence fails. Feedback channels mapped to HCRS dimensions—e.g., “Too technical”, “Missing information”, “Poorly structured”—enable more interpretable model training and closer alignment with user perceptions.

\paragraph{Model Integration.}
Human feedback should inform training, not just evaluation. Strategies include multi-objective learning, auxiliary classifiers for HCRS dimensions, and prompt engineering for targeted qualities. Care is needed with RLAIF, as over-reliance on synthetic evaluators risks misalignment with human values in sensitive domains.

\paragraph{Future Directions.}
Future work will extend validation of HCRS beyond health communication, test participatory feedback pipelines at scale, and explore adaptive dimension weighting.

\paragraph{RLAIF and Human-Centered Alignment in Health Communication.}
Recent advances in RL-based model alignment, particularly Reinforcement Learning from AI Feedback (RLAIF~\cite{lee2024rlaifvsrlhfscaling}, offer a major leap in scalability and efficiency. By using large language models (LLMs) to generate preference labels, RLAIF reduces the cost and time of annotation by more than an order of magnitude, while achieving competitive—and sometimes superior—results compared to traditional RLHF on standard benchmarks such as win rate and harmlessness.

Yet in domains like health communication, where trust, contextual sensitivity, emotional nuance, and sociocultural fit are essential, RLAIF faces critical limitations. Its optimization targets remain generic (e.g., helpfulness, harmlessness) and do not inherently capture the multidimensional, relational qualities required for impactful health information. Moreover, by inheriting preferences from pre-trained LLMs, RLAIF risks amplifying existing biases and overlooking authentic user values—an especially acute risk in high-stakes public health contexts. Finally, while RLAIF advances technical alignment, it does not natively support participatory, user-facing feedback loops or adaptive objectives such as trustworthiness, cultural relevance, and actionability.

The HCRS framework directly addresses these gaps. It moves beyond surface-level alignment toward user-driven, context-sensitive evaluation, integrates structured human feedback with participatory design, and operationalizes dimensions that matter in practice—clarity, trustworthiness, tone, cultural fit, and informed action—none of which current RLAIF pipelines explicitly model. In this way, HCRS can serve as both a complementary evaluation layer and a calibration signal for RLAIF-based training in sensitive domains.

\section{Conclusion}

This work introduced the Human-Centered Readability Score (HCRS), a five-dimensional framework for evaluating simplified health texts beyond surface features. By addressing clarity, trustworthiness, tone, cultural relevance, and actionability, HCRS fills critical gaps in current metrics and aligns evaluation with real-world user needs. We examined how existing metrics relate to these dimensions (Q1), proposed a composite score integrating automatic and human feedback (Q2), and outlined mechanisms to embed participatory, interactive evaluation into NLP workflows (Q3). Our empirical protocol sets the stage for validating HCRS across diverse populations and domains, enabling models that are both technically robust and socially responsive. 

While designed for public health communication, the approach extends to any domain where clarity, trust, and usability are paramount. Importantly, as discussed in Section~\ref{sec:planned-eval}, HCRS can also serve as a complementary evaluation layer and a calibration signal for RLAIF-based training pipelines in sensitive domains—helping to align scalable RL methods with nuanced, human-centered objectives.

\section{Limitations and Future Work}
While the proposed HCRS framework outlines a path toward more human-centered evaluation, this work has several limitations. First, the framework has not yet been empirically validated on large-scale, diverse user populations, so its generalizability remains to be tested. A key limitation of this work is therefore the lack of empirical validation. While we outline protocols and scenarios for participatory evaluation, we were not able to conduct a pilot study within the scope of this paper. Future work will therefore prioritize small-scale validation studies using micro-surveys and participatory workshops on real health communication materials. Second, the weighting of dimensions is currently conceptual and requires calibration against real-world user judgments. Third, operationalizing sociocultural and emotional dimensions relies on language resources and annotation protocols that may be domain- or language-specific. 

We also note that implementing all five dimensions in real evaluation settings may be resource-intensive. Some dimensions (e.g., clarity) can be partly automated, whereas others (e.g., trust or cultural relevance) require structured human feedback. Exploring hybrid setups that balance automation with targeted user input will be critical for feasibility.

Key directions for future evaluation, therefore, include: (i) Combining automatic metrics with validated user feedback instruments; (ii) Integrating participatory design into simplification evaluation; and (iii) Extending assessment to include sociocultural and emotional dimensions; and (iv) Exploring how HCRS can be integrated into existing NLP evaluation pipelines in a practical way, for example by combining automatic readability features with lightweight human feedback modules; (v) Conducting pilot user studies to provide proof-of-concept validation of the HCRS framework in practice. These directions will be essential for validating the HCRS framework in practice and expanding its applicability across domains where clarity, trust, and usability are critical.

While the HCRS framework was tailored for high-stakes public health communication, its core dimensions may require adaptation before use in other domains. Concepts such as trustworthiness or actionability are context-dependent and may need to be redefined based on the domain’s communicative norms and user expectations. Furthermore, calibrated human feedback, participatory evaluation, and relevant language resources would need to be retuned for new target populations. The framework should therefore be seen as a conceptual starting point, with significant work required to ensure generalizability, validity, and relevance outside of health contexts.

\section*{Acknowledgments}
This work was financially supported by the German Federal Ministry of Health (BMG) under grant No. ZMI5-2523GHP027. The project, titled “Strengthening National Immunization Technical Advisory Groups and Their Evidence-Based Decision-Making in the WHO European Region and Globally” (SENSE), is part of the Global Health Protection Programme (GHPP).


\bibliography{custom}

\end{document}